\documentclass{elsarticle} 
\usepackage{lineno}
\modulolinenumbers[5]
\usepackage[margin=2.5cm]{geometry}

\makeatletter
\def\ps@pprintTitle{%
 \let\@oddhead\@empty
 \let\@evenhead\@empty
 \def\@oddfoot{\centerline{\thepage}}%
 \let\@evenfoot\@oddfoot}
\makeatother

\journal{}
\usepackage{booktabs}
\usepackage{multirow}
\usepackage{algorithm}%
\usepackage{algpseudocode}
\usepackage{graphicx}
\usepackage{amssymb,mathrsfs}
\usepackage{amsmath,mathtools,eufrak}
\usepackage{color}
\usepackage{xspace}
\usepackage{algpseudocode}
\usepackage{bbm}
\usepackage{comment}
\usepackage{algorithmicx}
\usepackage{subfig}
\usepackage{psfrag}
\usepackage{lipsum}

\usepackage{tikz}
\usetikzlibrary{shapes,arrows}
\usetikzlibrary{positioning}
\usetikzlibrary{calc}
\usepackage[utf8]{inputenc}
\usepackage[english]{babel}
 
\tikzstyle{block}=[draw opacity=0.7,line width=1.4cm]
\definecolor{CranJ}{cmyk}{0,0.69,0.54,0.04} 
\definecolor{PinkJ}{cmyk}{0,0.71,0.43,0.12} 
\definecolor{Cran}{cmyk}{0,0.73,0.41,0.29} 
\definecolor{VRed}{cmyk}{0,0.75,0.25,0.2} 
\definecolor{ORed}{cmyk}{0,0.75,0.75,0} 
\definecolor{CBlue}{cmyk}{1,0.25,0,0} 

\newcommand{\real}{{\mathbb{R}}} \newcommand{\reals}{{\mathbb{R}}}

\newcommand{\argmin}{\operatorname{argmin}}

\newcommand{\vect}[1]{\boldsymbol{\mathbf{#1}}}
\newcommand{\vectsf}[1]{\vect{\mathsf{#1}}}

\newcommand{\oprocendsymbol}{\hbox{$\bullet$}}
\newcommand{\oprocend}{\relax\ifmmode\else\unskip\hfill\fi\oprocendsymbol}

\usetikzlibrary{quotes,arrows.meta}
\tikzset{
  annotated cuboid/.pic={
    \tikzset{%
      every edge quotes/.append style={midway, auto},
      /cuboid/.cd,
      #1
    }
    \draw [every edge/.append style={pic actions, densely dashed, opacity=.5}, pic actions]
    (0,0,0) coordinate (o) -- ++(-\cubescale*\cubex,0,0) coordinate (a) -- ++(0,-\cubescale*\cubey,0) coordinate (b) edge coordinate [pos=1] (g) ++(0,0,-\cubescale*\cubez)  -- ++(\cubescale*\cubex,0,0) coordinate (c) -- cycle
    (o) -- ++(0,0,-\cubescale*\cubez) coordinate (d) -- ++(0,-\cubescale*\cubey,0) coordinate (e) edge (g) -- (c) -- cycle
    (o) -- (a) -- ++(0,0,-\cubescale*\cubez) coordinate (f) edge (g) -- (d) -- cycle;
    \path [every edge/.append style={pic actions, |-|}]
    (b) +(0,-5pt) coordinate (b1) edge ["\cubex \cubeunits"'] (b1 -| c)
    (b) +(-5pt,0) coordinate (b2) edge ["\cubey \cubeunits"] (b2 |- a)
    (c) +(3.5pt,-3.5pt) coordinate (c2) edge ["\cubez \cubeunits"'] ([xshift=3.5pt,yshift=-3.5pt]e)
    ;
  },
  /cuboid/.search also={/tikz},
  /cuboid/.cd,
  width/.store in=\cubex,
  height/.store in=\cubey,
  depth/.store in=\cubez,
  units/.store in=\cubeunits,
  scale/.store in=\cubescale,
  width=10,
  height=10,
  depth=10,
  units=cm,
  scale=.1,
}

\begin{document}

\begin{frontmatter}

\title{ A Spatial-Temporal Decomposition Based Deep Neural Network for Time Series Forecasting}

\author{Reza Asadi\corref{mycorrespondingauthor}\fnref{myfootnote1}}
\author{Amelia Regan\fnref{myfootnote2}}
\fntext[myfootnote1]{Ph.D. candidate, Department of Computer Science, University of California Irvine}
\fntext[myfootnote2]{Professor of Computer Science, University of California Irvine}

\cortext[mycorrespondingauthor]{Corresponding author; rasadi@uci.edu}

\begin{abstract}
Spatial time series forecasting problems arise in a broad range of applications, such as environmental and transportation problems. These problems are challenging because of the existence of specific spatial, short-term and long-term patterns, and the curse of dimensionality. In this paper, we propose a deep neural network framework for large-scale spatial time series forecasting problems. We explicitly designed the neural network architecture for capturing various types of patterns. In preprocessing, a time series decomposition method is applied to separately feed short-term, long-term and spatial patterns into different components of a neural network. A fuzzy clustering method finds cluster of neighboring time series based on similarity of time series residuals; as they can be meaningful short-term patterns for spatial time series. In neural network architecture, each kernel of a multi-kernel convolution layer is applied to a cluster of time series to extract short-term features in neighboring areas. The output of convolution layer is concatenated by trends and followed by convolution-LSTM layer to capture long-term patterns in larger regional areas. To make a robust prediction when faced with missing data, an unsupervised pretrained denoising autoencoder reconstructs the output of the model in a fine-tuning step. The experimental results illustrate the model outperforms baseline and state of the art models in a traffic flow prediction dataset.
\end{abstract}


\end{frontmatter}

\section{Introduction and Literature Review}

Time series data arise in broad areas, such as engineering, medicine, finance, and economics. Various types of statistical and machine learning techniques have been applied on time series analysis. Recently, several new scalable time series analyses have been studied, such as forecasting \cite{lv2015traffic}, anomaly detection \cite{ahmad2017unsupervised}, classification \cite{zheng2014time} and clustering \cite{mikalsen2018time}. They illustrated the performance gains of these works over traditional time series techniques on large-scale problems. Moreover, spatial time series problems arise when there is a spatial dependency between neighboring time series. Spatial-temporal data arise in diverse areas of power grids \cite{bessa2015spatial}, load demand forecasting \cite{qiu2017empirical}, weather forecasting \cite{grover2015deep}, smart city applications \cite{tascikaraoglu2018evaluation}, and transportation systems, such as traffic flow forecasting \cite{polson2017deep}, \cite{zhang2017deep}.

Traffic flow prediction is one of the essential components of Intelligent Transportation Systems and one of the most challenging spatial-temporal problems, because of recurrent and non-recurrent patterns and the physical dynamics involved. Traffic flow prediction can assist travelers to make better decisions and improve traffic management, while decreasing traffic congestion and air pollution. Recently, smart devices increase the role of traffic flow prediction problems in our daily lives, which help people planning their travel and find the most efficient routes. With the advent of new sensing, computing and networking technologies, such as cameras, sensors, radars, inductive loops and GPS devices, a large volume of data are readily available \cite{zheng2016big}. These increasingly large data sets means that big data and techniques to handle these data plays a key role in the success of future transportation systems \cite{zhang2011data}. Hence, to improve the performance of transportation systems, researchers are motivated to take advantage of new spatial-temporal data-driven techniques and design scalable algorithms capable of processing large volume of data, such as deep neural networks , \cite{lv2015traffic}, \cite{al2015efficient}. 

\subsection{Background}

Starting in the 1970's with the original work of Gazis and Knapp \cite{gazis1971line} there have been many studies applying time series forecasting techniques to traffic flow prediction problem, including parametric techniques, such as auto-regressive
integrated moving average (ARIMA) \cite{kamarianakis2003forecasting} and Seasonal-ARIMA \cite{kumar2015short}, and statistical techniques, such as Bayesian analysis \cite{ghosh2007bayesian}, Markov chain \cite{yu2003short}
and Bayesian networks \cite{wang2014new}. However, there are several limitations on the models, because of prior assumptions, lack of handling missing data, noisy data, outliers, and the curse of dimensionality. Shallow architecture neural networks are capable of high dimension data, but cannot capture a high order computational complexity. With superior performance of deep neural networks on large-scale problems, they became an alternative technique applied on large-scale multi-variate time series forecasting problems.

Recently, there have been several attempts to design deep learning models for multi-variate time series forecasting problems. The primary work related to ours proposes a stacked autoencoder (SAE) model to learn traffic flow features and illustrate the advantage of SAE model versus Multi-layer Perceptron \cite{lv2015traffic}. In \cite{huang2014deep}, they propose stacked autoencoders with multi-task learning at the top layers of the neural network. A Deep Belief Network(DBN) composed by layers of restricted boltzman machine is proposed \cite{kuremoto2014time}. In \cite{wang2018optimal}, an ensemble of four categories of fully connected neural network is applied on time series forecasting problem. In \cite{qiu2014ensemble}, an ensemble of DBN with Support Vector Regression for aggregation of outputs is proposed for time series forecasting problem. However, in fully connected neural networks, the size increases exponentially with increasing input size, therefore the convergence of the model is computationally expensive and challenging. Several other neural network layers have been proposed to reduce the computational time and capture patterns in high order computationally complex temporal datasets.

Convolutional Neural Networks (CNN) extract features of various types of input data, such as images, videos, and audio. Weight sharing, the main feature of CNN, reduces the number of parameters in deep neural network models. These properties improve performance of learning algorithms by reducing complexity of parameters \cite{krizhevsky2012imagenet}. The performance of deep CNNs in multi-variate time series forecasting is examined; in \cite{ma2017learning}, a spatial-temporal relation of traffic flow data is represented as images. A CNN model is used to train from images and forecast speed in large transportation networks. In \cite{deng2018exploring}, they studied image-like representation of spatial time series data using convolution layers and ensemble learning. A convolution layer consider spatial structure in a euclidean space, which can miss some information on graph-structure data \cite{henaff2015deep}. As an alternative approach, following the work \cite{bruna2013spectral}, spatial dependency is captured using bi-directional diffusion convolutional recurrent networks \cite{li2017graph}. They illustrate a graph-structured representation of time series data capture spatial relation among time series. Moreover, in the presence of temporal data, recurrent neural networks have shown great performance in time series forecasting \cite{connor1994recurrent}. The vanishing gradient in deep Multi-layer perceptron and recurrent neural network problem is solved by employing a Long-Short Term Model (LSTM) \cite{sak2014long}, which significantly improves time series forecasting \cite{zhao2017lstm}, traffic speed prediction \cite{ma2015long} and traffic flow estimation with missing data \cite{tian2018lstm}. 


While convolutional neural networks can exhibit excellent performance on spatial data, and recurrent neural networks have advantages on problems with temporal data; spatial-temporal problems combine both of these. In \cite{xingjian2015convolutional}, they propose convolutional-LSTM layer for weather forecasting problem, in which consider spatio-temporal sequences. A convolutional deep learning model for multi-variate time series forecasting is proposed \cite{yi2017grouped}. They propose explicit grouping of input time series and implicit grouping using error back-propagation. In \cite{cheng2017deeptransport}, they use a CNN-LSTM model for downstream and upstream data to capture physical relationships among traffic flow data. A convolutional layer is followed by an LSTM layer for downstream and upstream traffic flow data. In \cite{liu2018short}, they illustrate a CNN and gated CNN followed by attention layers for spatial-temporal data. The capability of CNN-LSTM in learning spatial-temporal features are illustrated in above works. However, there is not any analysis on designing a neural network architecture with various components to separately capture spatial-temporal patterns.

\subsection{Contribution}

In the aforementioned works, spatial time series forecasting has been studied with the objective of proposing various types of convolution and recurrent neural network layers. However, spatial-temporal data have their specific patterns, which motivate us to use spatial and time series decomposition, and to explicitly consider various types of patterns in designing an efficient neural network architecture. There are some challenges in spatial-temporal data which should be considered in designing the deep neural network architecture. In spatial-temporal data, time series residuals are not only meaningless noise, but also related to physical properties and dynamical system of spatially dependent time series. Moreover, convolutional layers can capture spatial and short-term patterns, but sliding convolution kernels on spatial features miss network structure. In existence of long-term patterns, an LSTM layers shows great performance in forecasting problems because it can separately capture detrending data. Furthermore, a challenging problem is to address missing small spatial-temporal data in the time series forecasting problems.

In this paper, we address the problem of explicitly decomposing spatial-temporal patterns in designing a deep neural network and we illustrate its performance improvement on a large-scale traffic flow prediction problem. The contribution of the paper is described as follows:


\begin{itemize}
\item We illustrate an approach for explicitly considering various types of patterns in a deep neural network architecture for a spatial multi-variate time series forecasting problem.
\item We describe a Dynamic Time Warping-based clustering method and time series decomposition with the objective of finding compact regions with similar time series residuals.
\item A multi-kernel convolution layer is designed for spatial time series data, to keep the spatial structure of time series data and extract short-term and spatial patterns. It follows by a convolution-LSTM component to capture long-term patterns from trends, and a pretrained denoising autoencoder to have robust prediction to missing data.
\item The spatial and temporal patterns in traffic flow data is analyzed and the performance gains of the proposed model relative to baseline and state-of-art-the-art deep neural networks are illustrated for a traffic flow prediction, capturing meaningful time series residuals and a robust prediction to missing data.
\end{itemize}

The rest of the paper is as follows, in section II, we define the problem. In section III, the technical background of the proposed model are presented. In section IV, the proposed framework is illustrated, followed by the results of the work and conclusion discussed in section V.

\section{Problem Definition}
Time series data are a set of successive measurements, $\vectsf{X}_\mathsf{i} = \{\vectsf{x}_\mathsf{i}^1, \dots, \vectsf{x}_\mathsf{i}^\mathsf{\bar{t}} \}$, where $\vectsf{x}^\mathsf{t}_\mathsf{i}$ is an observed variable at location $\mathsf{i}$, time step $\mathsf{t}$ and $\mathsf{\bar{t}}$ is the size of the time series. A sensor $\mathsf{i}$ gathers $\vectsf{x}_\mathsf{i}^\mathsf{t}$ with corresponding $\mathsf{k}$ features $\vectsf{x}^\mathsf{t}_\mathsf{i} = \{\mathsf{x}_\mathsf{i}^{\mathsf{t},1} ,\dots, \mathsf{x}_\mathsf{i}^{\mathsf{t},\mathsf{k}}\}$. A spatial-temporal data is a set of $\mathsf{n}$ multi-variate time series $\vectsf{X} = \{\vectsf{x}_1, \dots, \vectsf{x}_\mathsf{n}$\}, represented by a matrix of $\vectsf{X} \in \reals^{\mathsf{n} \times \mathsf{\bar{t}} \times \mathsf{k}}$, where $\mathsf{n}$ is the number of sensors, which gather spatial-temporal data $\vectsf{X}$ in synchronous time  in a geographical area.

Given $\vectsf{X}$ as the set of all time series in a region, a spatial time series forecasting problem is cast as a regression problem. Given a time window of last $\mathsf{w}$ steps, and a horizon prediction $\mathsf{h}$, the objective is to predict  $\vectsf{X}_{\mathsf{output}}^\mathsf{t} = \{\vectsf{x}^{\mathsf{t}+1}, \dots,\vectsf{x}^{\mathsf{t}+\mathsf{h}}\}$, given $\vectsf{X}_{\mathsf{input}}^\mathsf{t} = \{  \vectsf{x}^{\mathsf{t}-\mathsf{w}}, \vectsf{x}^{\mathsf{t}-\mathsf{w}+1}, \dots, \vectsf{x}^{\mathsf{t}}\}$. The time window is used to only consider a small portion of previous temporal data for predicting horizon data, while we expect the model to memorize the long-term patterns. In equation (\ref{eq::ts_minimization}), an optimum parameter $\mathsf{\theta}^\star$ is the best model for forecasting time series data. In a neural network, $\mathsf{\theta}^\star$ is the weights of the model and the optimization algorithm minimizes the non-linear loss function $\mathsf{f}(., ., .)$ by solving following non-convex optimization problem,

    \begin{align}\label{eq::ts_minimization}
        \theta^\star = \argmin_{\mathsf{\theta}} \sum_{\mathsf{i}=1}^{\bar{\mathsf{t}}} \mathsf{f}(\vectsf{X}_{\text{input}}^\mathsf{i} , \vectsf{X}_{\text{output}}^\mathsf{i}, \mathsf{\theta})
    \end{align}

In this paper, given a spatial multi-variate time series data $\vectsf{X}$, a deep neural network predicts $\vectsf{X}_{\mathsf{output}} \in \reals^{\mathsf{n} \times \mathsf{h} \times \mathsf{\bar{k}}}$, where $\mathsf{\bar{k}}$ is the number of output features, for input data $\vectsf{X}_{\mathsf{input}} \in \reals^{\mathsf{n} \times \mathsf{w} \times \mathsf{k}}$. 



\section{Technical Background}
Here, we detail the core components of the proposed approach, including Fuzzy Hierarchical Agglomerative Clustering, Convolutional layers, Convolutional LSTM layers and Denoising Autoencoder.

\subsection{Dynamic Time Warping}\label{sec::DTW}

A Dynamic time warping (DTW) algorithm finds
an optimal path between two time series. It compares each point of first time series with second one. Hence, similar patterns occurred in different time slots are considered similar. A dynamic programming method finds the optimal match \cite{petitjean2012summarizing}. Here, we illustrate the DTW for $\mathsf{K}$-dimensional time series. Algorithm (\ref{alg:algo_dtw}) finds the minimum distance between two $\mathsf{K}$-dimensional time series with size of $\mathsf{N}$ and $\mathsf{M}$. 

\begin{algorithm}[t!]
  \caption{Multi-dimensional Dynamic Time Warping }\label{alg:algo_dtw}
  \begin{algorithmic}[1]
    \Procedure{$\delta$}{$\mathsf{a}$, $\mathsf{b}$}
        \State \textbf{return} $\sum_{\mathsf{k}=1}^\mathsf{K} |\mathsf{a}[\mathsf{k}] - \mathsf{b}[\mathsf{k}]|$
    \EndProcedure
    \Procedure{DTW}{$~\vectsf{X} = \{\vectsf{x}_1, \dots, \vectsf{x}_\mathsf{N} \}, \vectsf{Y} = \{\vectsf{y}_1, \dots, \vectsf{y}_\mathsf{M} \}$}\Comment{Two input time series}
    \State $\vectsf{X}$, $\vectsf{Y} \gets  \mathsf{Normalize}(\vectsf{X}$, $\vectsf{Y}$)
    \State $\vectsf{C}[1,1] \gets \delta(\vectsf{x}_1, \vectsf{y}_1)$ \Comment{Initialization of distance and path matrix}
    \State $\vectsf{P}[1,1] \gets (0, 0)$
    \For{ $\mathsf{i} \gets 2$ to $\mathsf{N}$ }
        \State~~~~$\vectsf{C}[\mathsf{i},1] \gets \vectsf{C}[\mathsf{i}-1,1] + \delta(\vectsf{x}_\mathsf{i}, \vectsf{y}_1)$ 
    \EndFor

    \For{ $\mathsf{j} \gets 2$ to $\mathsf{M}$ }
        \State~~~~$\vectsf{C}[1,\mathsf{j}] = \vectsf{C}[1,\mathsf{j}-1] + \delta(\vectsf{x}_1, \vectsf{y}_\mathsf{j}) $
    \EndFor
    \For{ $\mathsf{i} \gets 2$ to $\mathsf{N}$ }
	\For{ $\mathsf{j} \gets 2$ to $\mathsf{M}$ }
        \State~~~~$\vectsf{C}[\mathsf{i},\mathsf{j}] \gets \mathsf{min}(\vectsf{C}[\mathsf{i}-1,\mathsf{j}],\vectsf{C}[\mathsf{i},\mathsf{j}-1], \vectsf{C}[\mathsf{i}-1,\mathsf{j}-1]) + \delta(\vectsf{x}_\mathsf{i}, \vectsf{y}_\mathsf{j}) $
        \State~~~~$\vectsf{P}[\mathsf{i},\mathsf{j}] \gets \mathsf{minIndex}(\vectsf{C}[\mathsf{i}-1,\mathsf{j}],\vectsf{C}[\mathsf{i},\mathsf{j}-1], \vectsf{C}[\mathsf{i}-1,\mathsf{j}-1]) + \delta(\vectsf{x}_\mathsf{i}, \vectsf{y}_\mathsf{j}) $
         \EndFor
    \EndFor
      \State \textbf{return} $\mathsf{d} \gets \vectsf{C}[\mathsf{N}-1,\mathsf{M}-1]$, \Comment{Return the nonlinear distance of two time series}
    \EndProcedure 
  \end{algorithmic}
\end{algorithm}

\vspace{-0.2in}

\subsection{Fuzzy Hierarchical Clustering}\label{sec::FHC}

Given data points $\vectsf{X} = \{\vectsf{x}_1, \dots, \vectsf{x}_\mathsf{n} \}$, a fuzzy hierarchical clustering method finds a membership matrix $\vectsf{C} \in \reals^{\mathsf{n} \times \mathsf{c}}$, where $\mathsf{c}$ is the number of clusters and $\vectsf{C}_{\mathsf{i}\mathsf{j}} \in [0,1]$ illustrates the distance of data points $\mathsf{i}$ to cluster $\mathsf{j}$. 

To apply a DTW-based clustering method, the main challenge is to compute
the mean of a cluster addressed in \cite{gupta1996nonlinear}, \cite{niennattrakul2009shape}, \cite{petitjean2012summarizing}, because the initial values impacts on the final results of the algorithm. Hence, we consider fuzzy hierarchical clustering method without a need to find the cluster mean. Following the work of \cite{konkol2015fuzzy}, a complete-linkage is used for distances between clusters and a single-linkage is used for distance between points, and a point and a cluster. An algorithm (\ref{alg:algo_fhz}) finds the membership matrix of sensors to clusters.

The matrix $\vectsf{D}$ is the set of distances between all pair of time series and clusters, and it is initialized by all distances between points. The function $\mathsf{minDistance}(.)$ finds the closest pair of elements $(\mathsf{a},\mathsf{b})$ in the set $\vectsf{D}$, which $\mathsf{a}$ and $\mathsf{b}$ can be points or clusters. The matrix $\vectsf{C}$ is the list of clusters and their members. The function $\mathsf{updateClusters}(.,.,.)$ adds the selected pair of elements $(\mathsf{a},\mathsf{b})$ to the list of clusters. This function merges $\mathsf{a}$ and $\mathsf{b}$ into a new cluster. The matrix $\vectsf{A}$ is the list of assigned points, when a point is assigned into one cluster. Based on a new formation of clusters, the function $\mathsf{updateDistances}(.,.,.)$ finds the new distances between points and clusters. It updates the distance of all clusters and unassigned points to the new cluster. Moreover, it updates the fuzzy distance of all assigned points to the new cluster, and all points of new cluster to other clusters. The fuzzy distance between assigned point $\mathsf{u}$, and a cluster $\vectsf{c}_\mathsf{i}$ obtains by using equations (\ref{eq:fuzzy}).

\begin{align}\label{eq:fuzzy}
\mathsf{d}^{\text{min}}_\mathsf{u} &= \min \{ \mathsf{d}(\mathsf{u},\vectsf{c}_\mathsf{j}) | \vectsf{c}_\mathsf{j} \in \vectsf{C}\} \\
\mathsf{\mu}(\mathsf{u}, \vectsf{c}_\mathsf{i}) &=  \frac{\mathsf{d}^{\text{min}}_\mathsf{u}}{\mathsf{d}(\mathsf{u},\vectsf{c}_\mathsf{i}) + \mathsf{d}^{\text{min}}_\mathsf{u}},\\
\mathsf{d}(\mathsf{u},\vectsf{c}_\mathsf{i}) &= \min \{ (1 - \log_{\mathsf{m}}( \mu(\mathsf{u}, \vectsf{c}_\mathsf{i})))* \mathsf{d}(\mathsf{u},\vectsf{c}_\mathsf{i}), \mathsf{d}(\mathsf{u},\vectsf{c}_\mathsf{i}) \}
\end{align}

where $\mathsf{d}^{\text{min}}_\mathsf{u}$ is the minimum distance of a assigned point to any of the clusters, $\mathsf{\mu}(\mathsf{a},\mathsf{b})$ is membership value of assigned point $\mathsf{u}$ to the cluster $\vectsf{c}_\mathsf{i}$, $\mathsf{m}$ is a fuzziness parameter, and the distance function $\mathsf{d}(.,.)$ is based on single-linkage for each pair of points, or points and clusters, and complete-linkage for two clusters.


\begin{algorithm}[h]
  \caption{A DTW-based Fuzzy hierarchical Clustering on Spatial Time Series}\label{alg:algo_fhclustering}
  \label{alg:algo_fhz}
  \begin{algorithmic}
    \Procedure{FHC}{$\vectsf{X} = \{\vectsf{x}_1, \vectsf{x}_2, \dots, \vectsf{x}_n\}$}\Comment{$N$ input time series and spatial distances}
    \State {$\vectsf{D} = \{\}$;} \Comment{distance matrix}
    \State {$\vectsf{A} = \{\}$;}\Comment{List of assigned points}
    \State {$\vectsf{C} = \{\}$;} \Comment{List of clusters and their members}
    \State {$\vectsf{D} \gets \mathsf{initializeDistances}( \vectsf{X})$;}
    \While{Convergence is satisfied}
        \State $(\mathsf{a}, \mathsf{b}) \gets \mathsf{minDistance}(\vectsf{D})$;
        \State {$\vectsf{C}, \vectsf{A} \gets \mathsf{updateClusters}(\vectsf{C}, \vectsf{A}, \mathsf{a}, \mathsf{b})$}
        \State {$\vectsf{D} \gets \mathsf{updateDistances}( \vectsf{D}, \vectsf{A}, \vectsf{C}, \mathsf{a}, \mathsf{b})$;}
    \EndWhile\label{euclidendwhile}
    \State \textbf{return} $\vectsf{C}$\Comment{Return the list of clusters}
    \EndProcedure 
  \end{algorithmic}
\end{algorithm}





\subsection{Convolution Layer}\label{sec::Conv_layer}
A Convolutional layer uses three ideas of
local receptive fields, shared weights and spatial subsampling; making them effective and efficient models for exploiting the local stationary on grid data \cite{krizhevsky2012imagenet}. Given an input matrix $\vectsf{X} \in \reals^{\mathsf{n} \times \mathsf{\bar{t}} \times \mathsf{k}}$, a 2-dimension convolution layer has a weight matrix $\vectsf{W} \in \reals^{\mathsf{a} \times \mathsf{b} \times \mathsf{k}}$, called as a kernel, where $\mathsf{a} \leq \mathsf{n}$ and $\mathsf{b} \leq \mathsf{t}$. A convolution multiplication $\vectsf{X}*\vectsf{W}$ with strides $\mathsf{s}_1$ and $\mathsf{s}_2$ is obtained by sliding a kernel all over input matrix. The kernel is a shared weight which assume to have a locally stationary input data. Given $\vectsf{X}^\mathsf{l}$ as the input of layer $\mathsf{l}$, a layer $\mathsf{l}+1$ obtains by $\vectsf{X}^{\mathsf{l}+1} = \sigma(\vectsf{X}^\mathsf{l}*\vectsf{W}^\mathsf{l} + \vectsf{b})$ for an activation function $\sigma(.)$ and bias vector $\vectsf{b}$. Pooling layers $\vectsf{X}^{\mathsf{l}+1} = \mathsf{maxPool}(\vectsf{X}^\mathsf{l})$ among successive convolution layers reduces size of hidden layers, while extract features in locally connected layers, which selects the maximum value in a matrix of size $\vectsf{\bar{W}} \in \reals^{\mathsf{m} \times \mathsf{n}}$, and reduce the dimension of layers divided by $\mathsf{m}$ and $\mathsf{n}$.

\subsection{Convolution-LSTM layer}\label{sec::ConvLSTM_layer}
A Long-Short Term Memory (LSTM) is a special recurrent neural network cell with powerful modelling of long-term dependencies \cite{sak2014long}. A memory cell $\vectsf{c}^\mathsf{t}$, input gate $\vectsf{i}^\mathsf{t}$, output gate $\vectsf{o}^\mathsf{t}$ and forgot gate $\vectsf{f}^\mathsf{t}$ works together in hidden units $\vectsf{h}^t$. Given $*$ convolution operator and $\circ$ a Hadamard product, a convolution LSTM is as follows \cite{xingjian2015convolutional},

    \begin{align}
\vectsf{i}^\mathsf{t} &= \sigma(\vectsf{W}^{\vectsf{x}\mathsf{i}}*\vectsf{x}^\mathsf{t} + \vectsf{W}^{\vectsf{h}^\mathsf{l}}*\vectsf{h}^{\mathsf{t}-1} + \vectsf{W}^{\vectsf{c}_\mathsf{i}} \circ \vectsf{c}^{\mathsf{t}-1}+\vectsf{b}^\mathsf{i}) \\
\vectsf{f}^\mathsf{t} &= \sigma(\vectsf{W}^{\vectsf{x}\mathsf{f}}*\vectsf{x}^\mathsf{t} + \vectsf{W}^{\vectsf{h}_\mathsf{f}}*\vectsf{h}^{\mathsf{t}-1} + \vectsf{W}^{\mathsf{c}\mathsf{f}} \circ \vectsf{c}^{\mathsf{t}-1}+\vectsf{b}^\mathsf{f}) \\
\vectsf{c}^\mathsf{t} &= \vectsf{f}^\mathsf{t} \circ \vectsf{c}^{\mathsf{t}-1} + \vectsf{i}^\mathsf{t} \circ \tanh(\vectsf{W}^{\mathsf{x}\mathsf{c}}*\vectsf{x}^\mathsf{t} + \vectsf{W}^{\mathsf{h}\mathsf{c}}*\vectsf{h}^{\mathsf{t}-1}+\vectsf{b}^\mathsf{c}) \\
\vectsf{o}^\mathsf{t} &= \sigma(\vectsf{W}^{\mathsf{x}\mathsf{o}}*\vectsf{x}^\mathsf{t} + \vectsf{W}^{\mathsf{h}\mathsf{o}}*\vectsf{h}^{\mathsf{t}-1} + \vectsf{W}^{\mathsf{c}\mathsf{o}} \circ \vectsf{c}^{\mathsf{t}}  +\vectsf{b}^\mathsf{o}) \\
\vectsf{h}^\mathsf{t} &= \vectsf{o}^\mathsf{t} \circ \tanh(\vectsf{c}^\mathsf{t})
    \end{align}

A convolution-LSTM layer have same structure of convolution layers, but having LSTM cells. The gates prevent a gradient from vanishing quickly by storing it in memory. The convolution-LSTM layer has a input of $\vectsf{X} \in \reals^{\mathsf{w} \times \mathsf{a} \times \mathsf{b} \times \mathsf{k}}$, where $\mathsf{w}$ is time windows and the matrix $\vectsf{W} \in \reals^{\mathsf{a} \times \mathsf{b} \times \mathsf{k}}$ is the spatial information on a grid of size $\mathsf{a}$ and $\mathsf{b}$ and each element $\vectsf{W}_{\mathsf{i}\mathsf{j}}$ has $\mathsf{k}$ features.

\subsection{Denoising Stacked Autoencoder}\label{sec::DAED}
Given an input $\mathsf{m}$-dimension data $\vectsf{x} \in \reals^\mathsf{m}$, an autoencoders transforms input with a non-linear function $\vectsf{h} = \sigma(\vectsf{x}\vectsf{W}^1+\vectsf{b}^1), \vectsf{h} \in \reals^\mathsf{d}$, where $\mathsf{d} < \mathsf{m}$ is lower dimension space \cite{vincent2008extracting}. The decoder generates $\vectsf{z} = \sigma(\vectsf{h}\vectsf{W}^2+\vectsf{b}^2)$, where $\vectsf{z} \in \reals^\mathsf{m}$. In the training process the objective is to reconstruct $\vectsf{x}$, by minimizing loss function $\mathsf{L}(.,.,.)$, such as least square function, between $\vectsf{x}$ and $\vectsf{z}$ and obtaining optimum model parameters $\vectsf{\theta}^\star$ for all $\mathsf{m}$ input data as follows,

\begin{align}
	\vectsf{\theta}^\star = \argmin_{\vectsf{\theta}} \sum_{\mathsf{i}=1}^{\mathsf{m}} \mathsf{L}(\vectsf{x}_\mathsf{i} ,\vectsf{z}_\mathsf{i}, \vectsf{\theta})
\end{align}

Stacked autoencoders $\mathsf{\bar{f}}(.)$ are a set of multiple  autoencoder layers, in which the input of each layer is the output of previous layer \cite{vincent2010stacked}. The input data is corrupted with some noise, while the output remains unchanged. Adding noise to the input data and training the neural network to reconstruct the clean output, helps the neural network to learn robust features to noisy data. The noise can be added to the network using a dropout function in each layer $\vectsf{h}^{\mathsf{l}+1} = \mathsf{dropout} (~\sigma(~\vectsf{h}^{\mathsf{l}}\vectsf{W}^\mathsf{l}+\vectsf{b}^\mathsf{l}), \alpha)$, in which randomly $\alpha$ percent of neurons are dropped in each round of training process. Unsupervised training of stacked autoencoders with the form of $\vectsf{x} = \mathsf{\bar{f}}(\vectsf{x})$ is capable of reconstructing the original data in the presence of noisy or missing data  \cite{zhou2017delta}, \cite{vincent2010stacked}, \cite{gondara2018mida}.

\section{Methodology:}\label{eq::dnnf}
\begin{figure}[t] 
\hspace{-0.5cm}\centering
\includegraphics[width=13cm]
{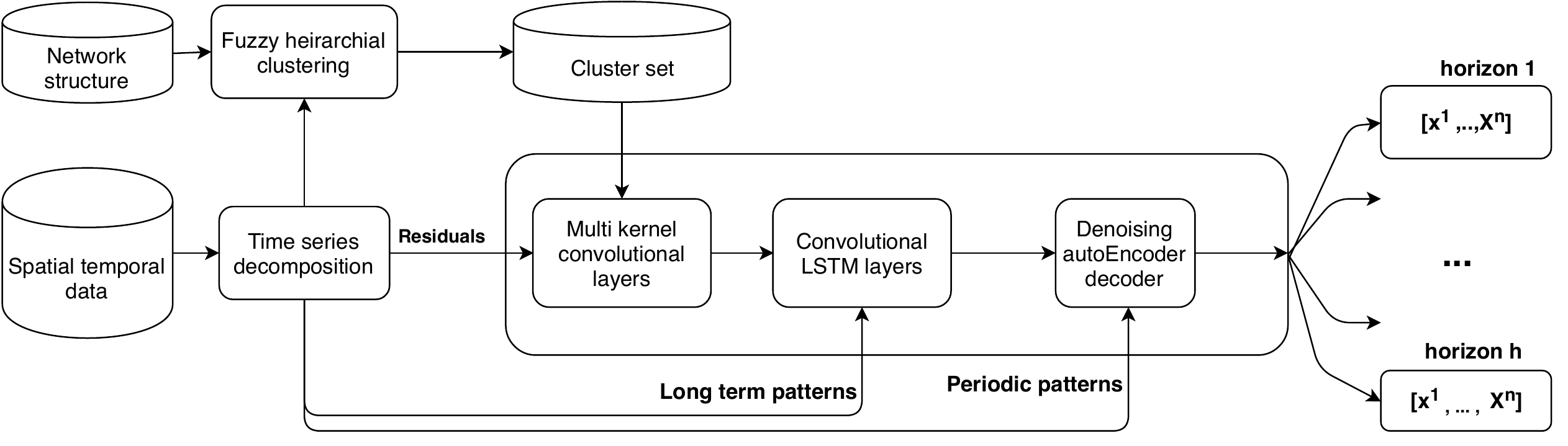}
\caption{The proposed framework for spatial multi-variate time series forecasting problem}\label{fig::methodology_1}
\vspace{-0.1in}
\hspace{-0.2in}
\end{figure}

In this section, we describe the architecture of the proposed deep learning framework for spatial time series forecasting problem. The proposed framework is illustrated in Fig. (\ref{fig::methodology_1}). The network structure represents the distance between neighboring sensors, and the spatial-temporal data includes time series data for each sensor. 

\subsection{Preprocessing}
A time series decomposition method is applied on input time series $\vectsf{X} \in \real^{\mathsf{n} \times \mathsf{w} \times \mathsf{k}}$, which generates three time series components of $\vectsf{X} = (\vectsf{S}, \vectsf{T}, \vectsf{R})$, which are seasonal, trends and residuals of time series, respectively. In spatial time series data, residuals can be different than only noise. For example, in a transportation network, time series residuals can be caused by the traffic evolution of the transportation network and they are meaningful patterns among neighboring time series, analyzed in section \ref{sec::exp_ana} experimental results. 

To apply algorithm (\ref{alg:algo_fhclustering}) on time series residuals, we consider a set $\mathsf{G}$ for geographically nearest neighbors of sensors. The algorithm updates single-linkage distances between two time series from set $\mathsf{G}$; thus the clusters would not be distributed in a geographical area. Since some of the sensors might affect more than one regional cluster, the output of clustering algorithm finds fuzzy membership of each sensor to their similar clusters. Each sensor $\vectsf{x}_\mathsf{i}$ has a membership to some cluster $\vectsf{c}_\mathsf{j} \in \vectsf{C}$. We say two time series $\vectsf{x}_\mathsf{i}$ and $\vectsf{x}_\mathsf{j}$ are similar, if two time series have similar patterns over some time shift, or have zero distance from each other. Hence, for a given distance function $\tau(.,.)$, which we consider DTW, a fuzzy hierarchical clustering algorithm finds the cluster of sensors with similar residual time series by finding the clusters in which the distance of its members is minimized. To represent short-term similarity among neighboring time series, we used a rolling window on training data and getting average of corresponding DTW distances. A rolling window finds similarity between short-term time window of neighboring area. To reduce computational time, rolling window is only applied, when there are high interaction among neighboring time series. For example, in traffic flow data, the interaction among neighboring sensors increases peak hours and congestion time periods. Applying algorithm \ref{alg:algo_fhclustering} with aforementioned modifications on spatial time series finds fuzzy clusters of time series based on DTW distance.

\subsection{Neural Network Architecture}

\begin{figure}[t] 
\hspace*{-0.5cm}\centering
\includegraphics[width=14
cm]{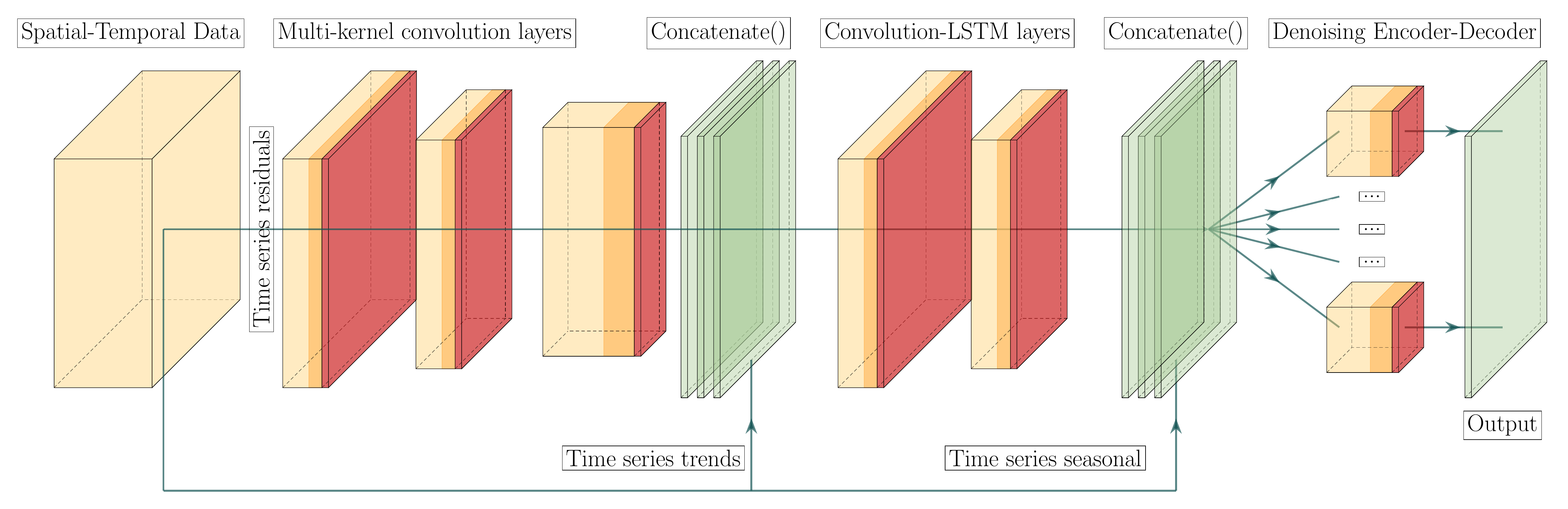}\caption{The proposed spatial-temporal decomposition deep neural network architecture}\label{fig::methodology_2}\vspace{-0.1in}\hspace{-0.2in}
\end{figure}

The details of the deep neural network is represented in Fig. \ref{fig::methodology_2}. Time series residuals are the first input of the neural network, detrended and represented with a matrix of $\vectsf{X} \in \reals^{\mathsf{s} \times \mathsf{w} \times \mathsf{k}}$. A convolutional component is applied to extract patterns from time series residuals. For a given set of time series $\vectsf{X}$, a general convolution kernel slides on first and second axis. However, because the sensors can have a spatial structure, like sensors in a transportation network, sliding a kernel on sensors cannot keep the structure of the network. Moreover, each sensor's time series residuals are only dependent to small regions in the network. Hence, we propose a multi-kernel convolution layer, which receives the cluster set and residual time series data. A kernel $\vectsf{W}_\mathsf{i}$ for a cluster $\mathsf{i}$, is described with weight matrix $\vectsf{W}_\mathsf{i}$ where $\vectsf{W}_{\mathsf{i}\mathsf{j}} \neq 0$, if $\mathsf{j} \in \mathsf{c}_i$. In other words, the size of trainable variables for a kernel, corresponding to cluster $\mathsf{i}$, is $\vectsf{W}_\mathsf{i} \in \reals^{|\vectsf{C}_\mathsf{i}|, \mathsf{w}, \mathsf{k}}$. Only the sensors in cluster $\mathsf{i}$, has a local connectivity to same residual time series. The kernel slides over time axis and obtain hidden units $\vectsf{h}_\mathsf{i} = \mathsf{pool}(~\sigma(~\vectsf{R}_\mathsf{i}^\mathsf{t} \vectsf{W}_\mathsf{i} + \vectsf{b}_\mathsf{i}))$ for all $ \mathsf{i} \in \{1, \dots, |\vectsf{C}|\}$, where $\mathsf{pool}$ is pooling layer. Several convolution-RELU-Pooling layers extracts short-term and spatial patterns from the time series residuals in each neighborhood. The output of kernels are concatenated and connected to a fully-connected layer $\vectsf{h}^{\mathsf{l}+1} = \mathsf{concat}(\mathsf{FC}(\mathsf{h}_1^\mathsf{l}),\dots, \mathsf{FC}(\mathsf{h}_{|\vectsf{C}|}^\mathsf{l}))$ and represented with a hidden layer $\vectsf{h}^{\mathsf{l}+1} \in \reals^{\mathsf{w} \times \mathsf{s} \times \mathsf{v} \times 1}$, where $\mathsf{v}$ is the number of represented features in convolution layers and $\mathsf{s}$ is the total number of sensors.

The time series trends represent long-term patterns. The trends of time series $\vectsf{T}$ concatenate to $\vectsf{h}^{\mathsf{l}+1}$ on the last axis, $\vectsf{h}^{\mathsf{l}+2} = \mathsf{concat}(\vectsf{h}^{\mathsf{l}+1}, \mathsf{FC}(\vectsf{T}), \mathsf{axis} = 4)$  which results in $\vectsf{h}^{\mathsf{l}+2} \in \reals^{\mathsf{w} \times \mathsf{s} \times \mathsf{v} \times 2}$. Unlike residuals which represent physical dynamics of the problem and there is only similarity between neighboring areas, trends can represent global changes in the spatial-temporal data. Hence, we consider LSTM cells to capture long-term patterns for the concatenated output of the extracted features of smaller regions. The model follows by a 2-dimension convolution LSTM layers. A 2-dimension convolution LSTM layer, described in section \ref{sec::ConvLSTM_layer}, receives an input $\vectsf{h}^{\mathsf{l}+2}$, and apply the convolution on the matrix of size $(\mathsf{a}=\mathsf{s}, \mathsf{b} \leq \mathsf{v})$ with two channels. This convolutional layer has different architecture with the first multi-kernel convolution layer, that is, each neural cell is an LSTM cell and is applied on all input sensors. Some layers of convolution LSTM layers extract features from residuals and trends. Seasonal patterns represents repeated patterns for the given time horizon. The output is concatenated with seasonal patterns of time window $\{\mathsf{t}-\mathsf{w}, \dots, \mathsf{t}+\mathsf{h}\}$.  It follows by a fully-connected layer. The output is $\vectsf{\bar{y}} \in \reals^{\mathsf{s} \times \mathsf{h} \times \mathsf{k}}$, where $\mathsf{h}$ is prediction horizon. The output $\vectsf{\bar{y}}$ consists of predicted values for all sensors in prediction horizon.

One of the challenges in spatial-temporal data is to have a robust prediction in existence of missing, noisy data. Hence, we consider an autoencoder layer in as the last component of the model. A denoising autoencoder decoder reconstructs the last output $\vectsf{\bar{y}}$ for each cluster. In the pretraining step, for a prediction horizon $\mathsf{h}$ and a cluster $\mathsf{j}$, each denoising autoencoder decoder generates $\vectsf{x} = \mathsf{DA}_{\mathsf{j}}(\vectsf{x})$, where $\vectsf{x} \in \reals^{\mathsf{s} \times \mathsf{h} \times \mathsf{k}}$ and drop out layer are between each successive layers. A denoising autoencoder component generates a predictions $\vectsf{\bar{y}}_\mathsf{d} = \mathsf{DA}(\vectsf{\bar{y}})$. As the output of autoencoders is designed based on the clusters, there are some sensors $\vectsf{x}_{\mathsf{k}} \in \vectsf{c}_\mathsf{i} \cap \vectsf{c}_\mathsf{j}, \mathsf{i} \neq \mathsf{j}$, where the fully-connected target layer $\mathsf{FC}^\mathsf{t}$ is connected to all common variables between denoising autoencoders with a linear activation function  $\vectsf{y}_{\mathsf{output}} = \mathsf{FC}^\mathsf{t}(\mathsf{DA}_1(\vectsf{\bar{y}}) , \dots, \mathsf{DA}_\mathsf{|C|}(\vectsf{\bar{y}}))$. In the training process the objective is to minimize loss function $\mathsf{L}(.,.,.)$, such as mean square error function, between $\vectsf{y}_{\mathsf{output}}$ and real values $\vectsf{y}$, and obtaining optimum model parameters $\vectsf{\theta}^\star$ for input data using stochastic gradient descent,

\begin{align}
	\vectsf{\theta}^\star = \argmin_{\vectsf{\theta}} \sum_{\mathsf{i} = 1}^{|\vectsf{X}|} \mathsf{L}(\vectsf{x}_\mathsf{i} ,\vectsf{y}_\mathsf{i}, \vectsf{\theta})
\end{align}

\section{Experimental Analysis}\label{sec::exp_ana}
We illustrate the analysis and the performance of the proposed methodology on traffic flow data.

\subsection{Data Set}
\begin{figure}[h]
\hspace*{-0.5cm}\centering
\includegraphics[width=8cm]{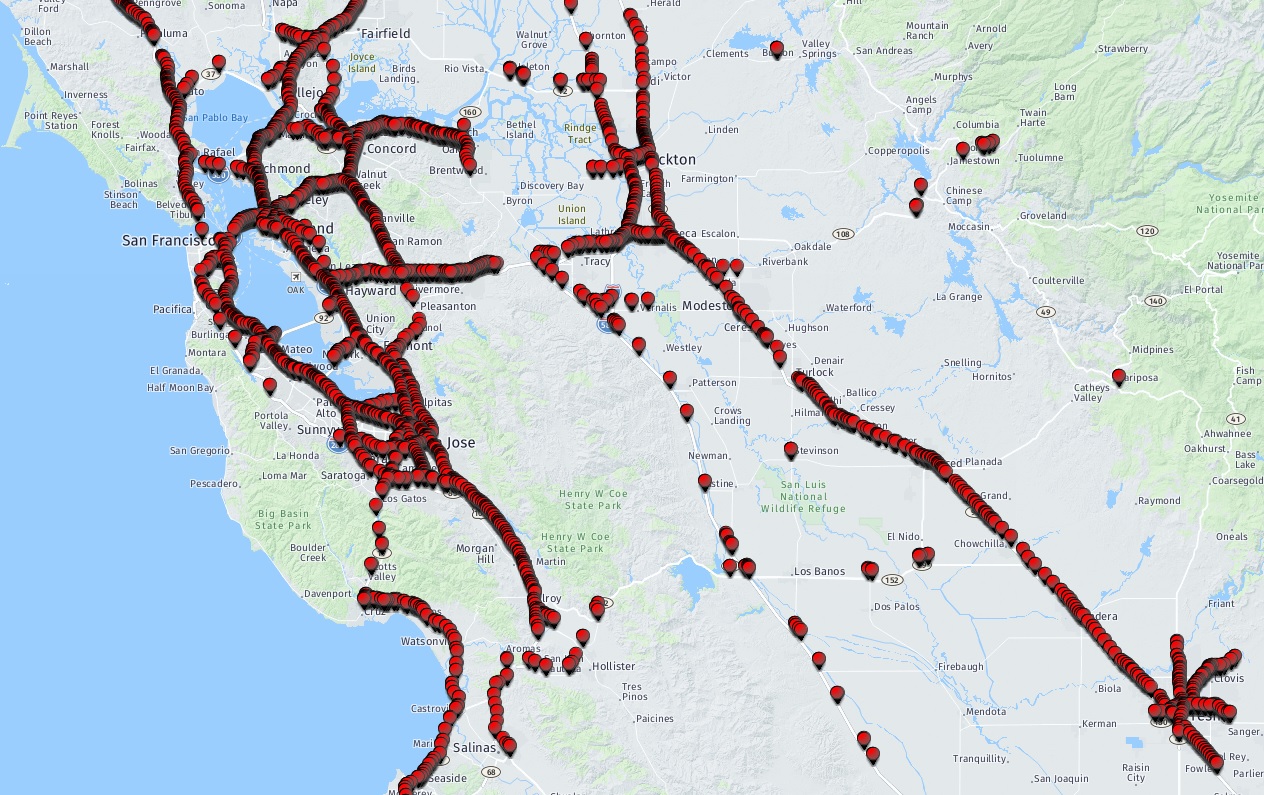}\caption{Traffic sensors over a Bay Area, California. The red dots represents loop detector sensors on highways.}\label{fig::network}\vspace{-0.1in}
\end{figure}

We use traffic flow data from the Bay Area of California represented in Fig. \ref{fig::network} which is commonly used and available in PEMS \cite{californiapems}. The traffic flow has been gathered every 30 seconds and aggregated every 5 minutes in the dataset. Each sensor on highways of California has flow, occupancy and speed at a time stamp. A sensor is a loop detector device in mainline, off-ramp or on-ramps locations. In preprocessing, we selected 597 sensors which they have more than $90\%$ observed values in the first six months of 2016.

\subsection{Pattern analysis in traffic data}
To illustrate the specific characteristics of traffic data arise from dynamic of traffic flow, we analyze the spatial, short-term and long-term patterns.

\begin{figure}[h]\hspace{-0.0in}
  \centering   
  \subfloat[The observed flow data.]{
    \includegraphics[height=1.2in]{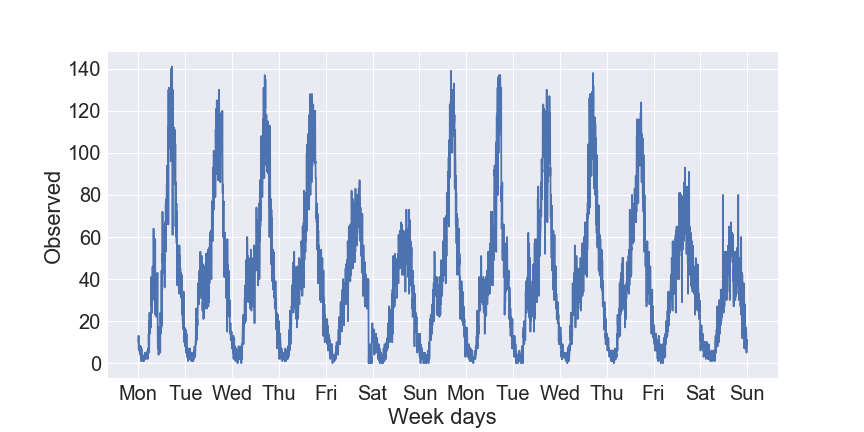}
  }~~~
  \subfloat[Seasonal patterns traffic  flow data]
  {
    \includegraphics[height=1.2in]{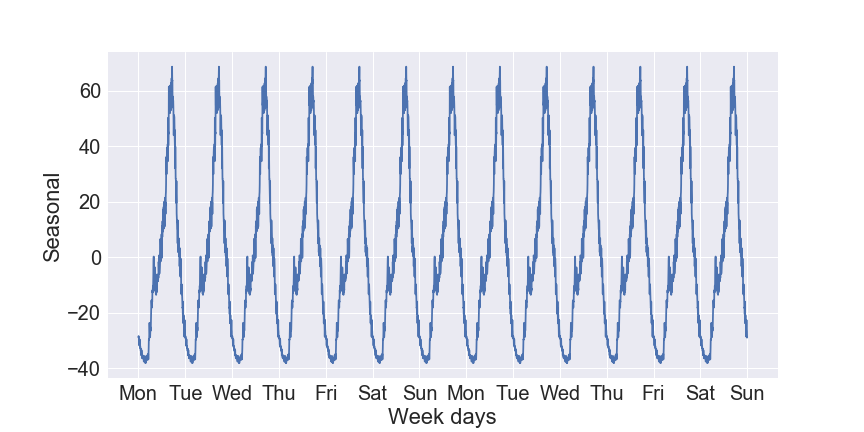}
  }\\ 
  \hspace{-0.0in}
  \subfloat[Trends of traffic flow data]{
    \includegraphics[height=1.2in]{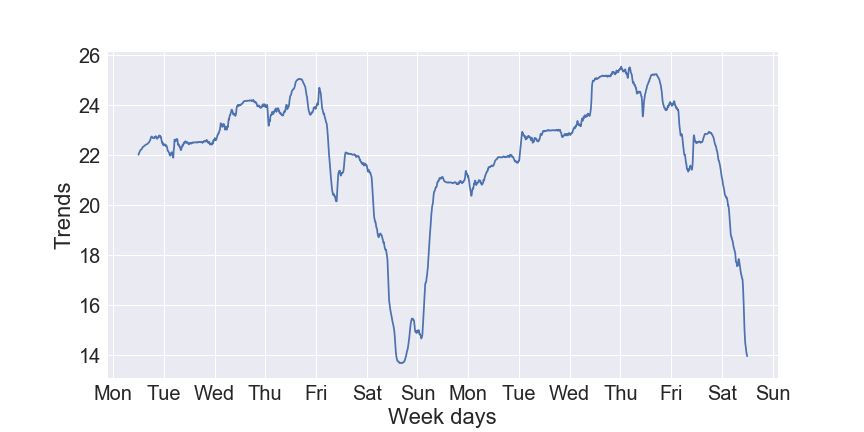}
  }~~~
  \subfloat[Residuals of traffic flow data]
  {
    \includegraphics[height=1.2in]{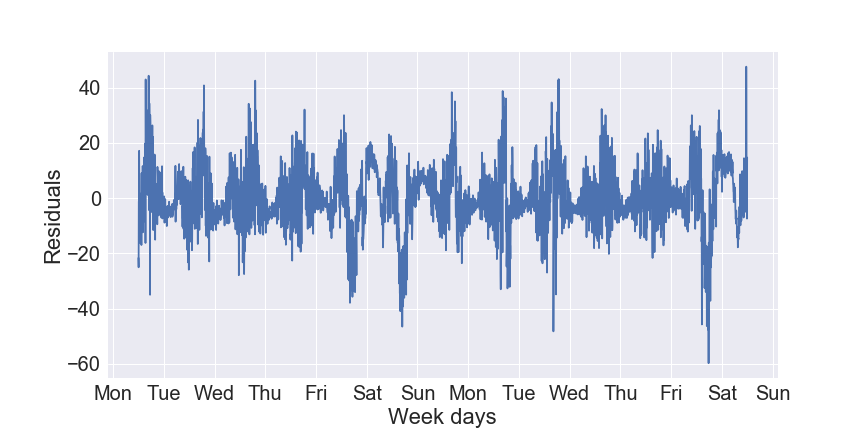}
  }\\
  \caption{Time series decomposition of traffic flow data for one station in the network with daily frequency.}\label{fig::ts_decomp}\vspace{-0.1in}
\end{figure}

In Fig. \ref{fig::ts_decomp}, an additive time series decomposition of traffic flow data is illustrated for one station. Given a one day frequency, time  series decomposition has similar, repeated (seasonal) daily patterns. Moreover, there are long-term weekly patterns, shown as trends $\vectsf{T}$. The long-term patterns, such as seasonal and trends, arise from similar periodic patterns, generated outside of the highway network. In other words, they are related to origin-destination matrix of vehicles in the network.
The residual patterns are not only random noise, but also the results of spatial and short-term patterns and related to the evolution of traffic flow or sudden changes in smaller regions of the network. Since they have more similar patterns for neighboring time series, illustrated in section \ref{sec::resid_pattern}.

\subsubsection{Residual time series in traffic flow data}\label{sec::resid_pattern}

\begin{figure}[h!]
\centering
\includegraphics[width=10
cm]{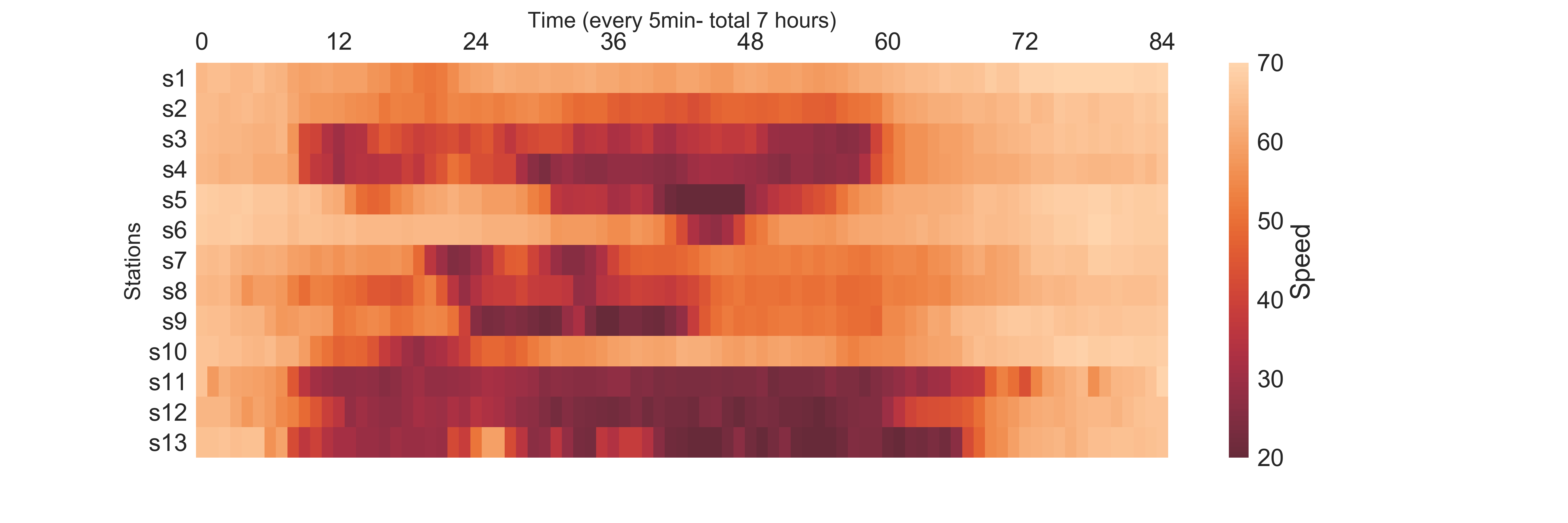}\caption{Image representation of a speed value of 13 successive sensors over 7 hours in a highway. It shows neighboring sensors have same congestion hours.}\label{fig::ts_image}
\end{figure}

A time series decomposition consists of residuals, trends and seasonal components. The residuals are interpreted as random noise for time series data. However, in traffic flow data, the residuals are the results of physical evolution of the network. In non data-driven traffic flow problems, first-order and second-order traffic flow fundamental diagram shows the relation between traffic flow, occupancy and speed. Given a wave-speed $\mathsf{w}_\mathsf{i}$, free speed $\mathsf{s}_\mathsf{i}$ and maximum density $\mathsf{b}_\mathsf{i}$ in a road segment $\mathsf{i}$, the first order dynamical traffic flow theorem approximates flow by $\mathsf{f}_\mathsf{i}(\mathsf{o}_\mathsf{i}) = \min\{\mathsf{s}_\mathsf{i} \mathsf{o}_\mathsf{i}, \mathsf{w}_\mathsf{i}(\mathsf{b}_\mathsf{i}- \mathsf{o}_\mathsf{i})\}$. Wave-speed reduces flow in high occupancy. In Fig. \ref{fig::ts_osf}, we examine the non-linear relation of flow, speed and occupancy in one day and one station. It illustrates the relation between high occupancy and reduction of average speed in the road segment, leads to traffic congestion. This property of fundamental traffic flow diagram illustrates non repeated, residual patterns in traffic flow data as a result of congestion. The congestion propagation in a transportation network illustrates the relation among neighboring sensors in highways, described in Fig. \ref{fig::ts_osf} for flow data of three successive sensors. Congestion propagates over this sensors with nearly 20 min delay. For a larger area, in Fig. \ref{fig::ts_image}, the speed of 13 successive sensors is represented in an image. The reduction of speed in peak hours is presented with darker colors. It illustrates the reduction in speed is similar in neighboring areas.



\begin{figure}[h!]
  \centering   
  \subfloat[An example of a relation among flow, occupancy and speed. Occupancy, with value more than 8, decreases average speed as a result of wave-speed.]{
    \includegraphics[height=1.6in]{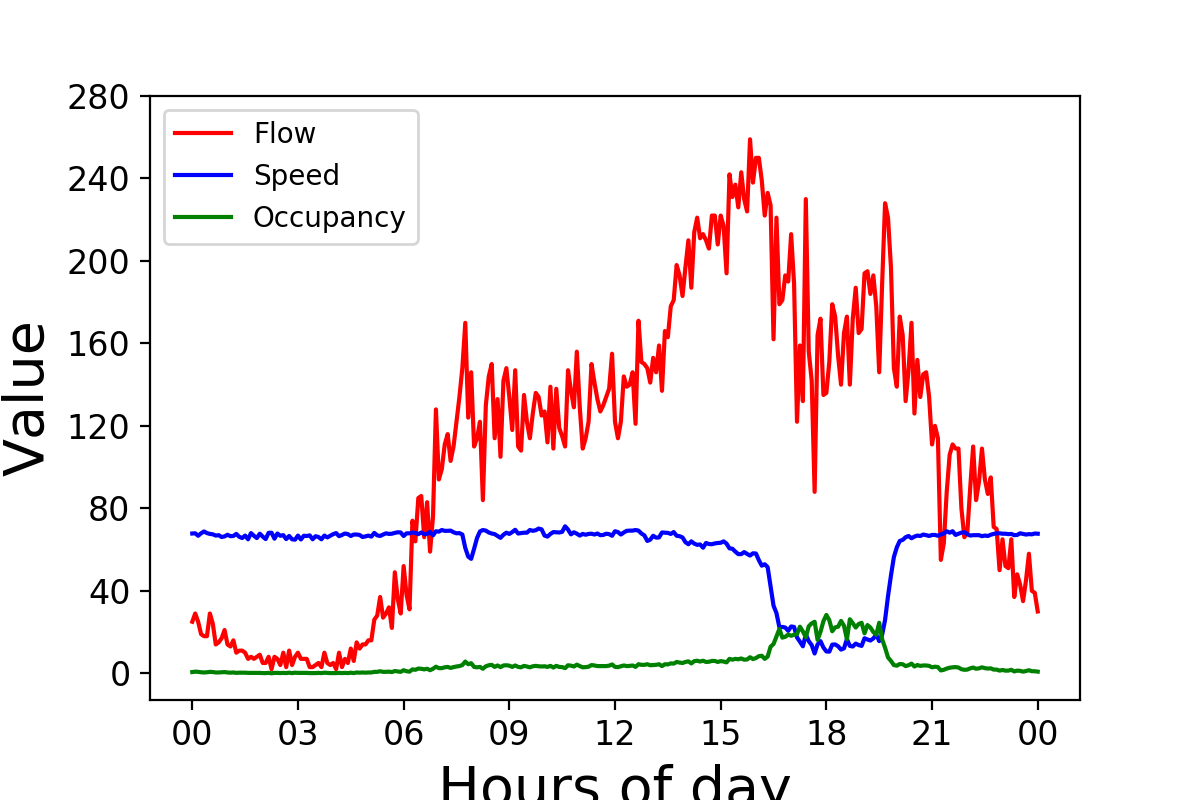}
  }~~
  \subfloat[Log plot to represent the linear relation between occupancy and flow with free-speed near to 70.]
  {
    \includegraphics[height=1.6in]{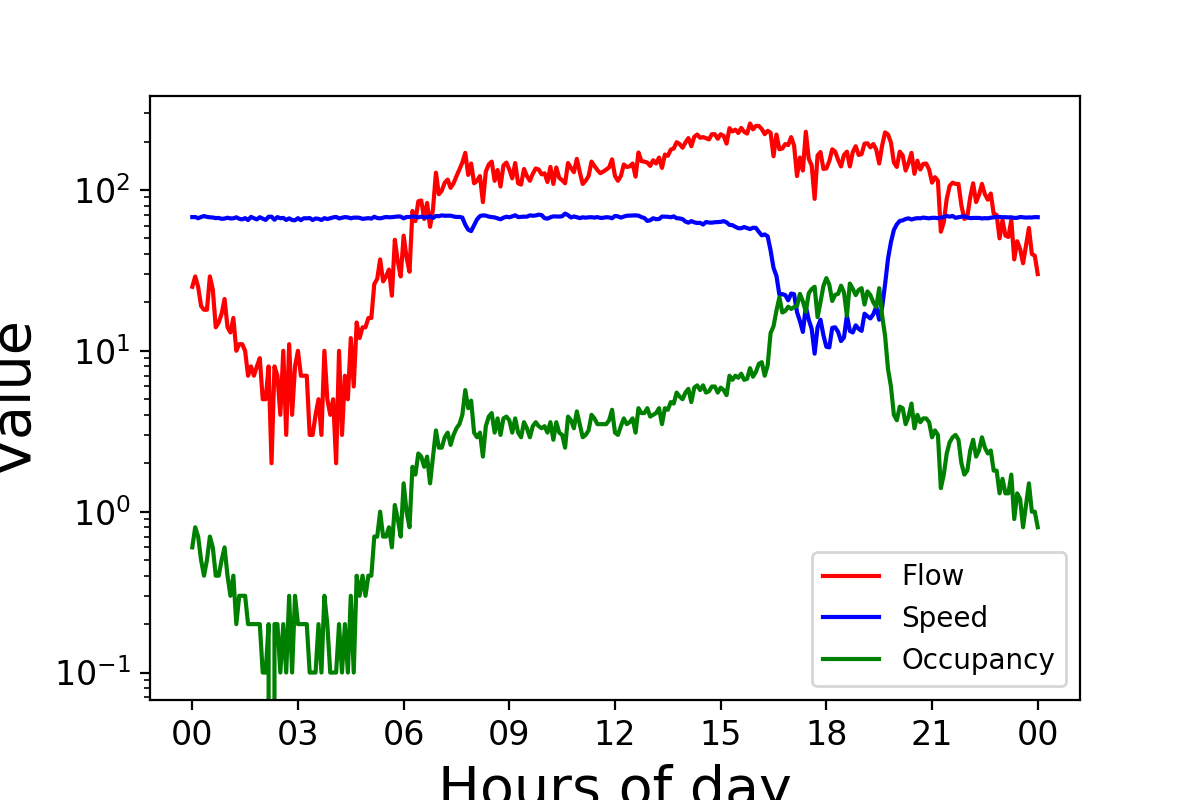}
  }\\ 
  \caption{The relation between Occupancy, Speed and Flow}\label{fig::ts_osf}\vspace{-0.1in}
\end{figure}

\begin{figure}[h!]
  \centering   
  \subfloat[The upstream and downstream of sensors spatially affect each other.]{
    \includegraphics[height=0.9in]{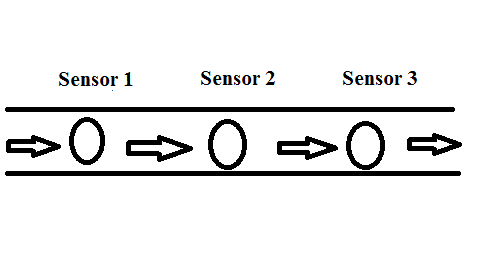}
  }~~
  \subfloat[The reduction in speed of sensor 1 and 2 twice happens in this plot, which there is 20 minute delay in congestion propagation.]
  {
    \includegraphics[height=1.9in]{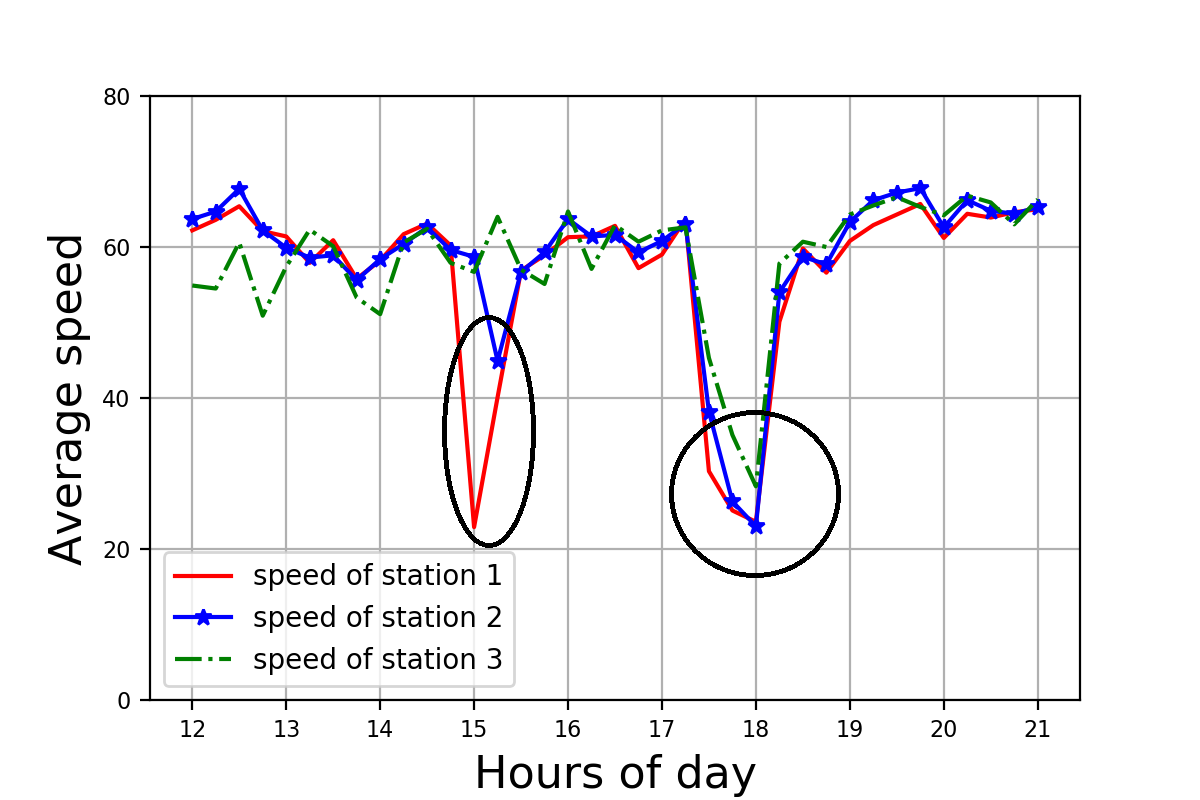}
  }\\ 
  \caption{The congestion propagation in successive sensors.}\label{fig::ts_osf}
\end{figure}

\subsection{Fuzzy Hierarchical Clustering}\label{eq::fhc}

\begin{figure}[h!]
\hspace*{-2cm}\centering
\includegraphics[width=6
cm]{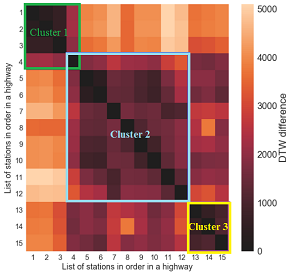}\caption{The table shows the Dynamic Time Warping distance distance of time series residuals among 15 stations on a highway. The result of hierarchical clustering method is illustrated with three clusters. The distance values near to diagonal have lower distance.}\label{fig::clustering_ts}\vspace{-0.1in}\hspace{-0.2in}
\end{figure}

In this section, we illustrate the results of fuzzy clustering applied on time series residuals. In Fig. \ref{fig::clustering_ts}, the DTW distance matrix shows residual similarity among neighboring sensors. The matrix shows the average DTW distance for peak times of training data, which has the highest dependency because of high values of occupancy. Each cluster would be obtained by comparing neighboring sensors. On the elements near to diagonal, the lowest distance values shows the similarity between neighboring values.

After preprocessing of time series data, there are 597 sensors with complete data over a period of six months. The fuzzy clustering finds the membership of each sensor to clusters. In the fuzzy membership matrix, we consider threshold of 0.1. All the sensors which has a membership value of more than 0.1, they would be considered as the members of clusters. We also consider the average size of clusters to be less than 10 miles. The agglomerative clustering stops when the average become greater than 10. As the clustering is applied on mainline stations, we also added the on-ramp and off-ramp sensors to the closest mainline stations. The result of fuzzy heirarchial clustering method, has 64 clusters, where the average number of elements is 9.7 with standard deviation of 4.2 and minimum cluster size of 3 and maximum of 14. The length of smallest and largest cluster is 0.3 mile and 32.1 mile. And there are 53 sensors which appear in more than one cluster, nearly $10\%$ of total sensors. To examine the relation between trends in one spatial area, using a rolling window, we obtain DTW distance of each pair of sensors. For a time window, we normalized trends, by subtracting all time stamp values from the last value. The average DTW distance is 0.7, for all pairs of sensors, which shows the high similarity of trends. By contrast, presented in section \ref{sec::resid_pattern}, the average DTW distance of time series residuals for all pair of sensors is 4.5, while applying the fuzzy clustering method on time series reduces the average DTW of clusters to 0.6. As a result, we only apply fuzzy clustering on time series residuals.

\subsection{Results of comparison}
Our model is demonstrated to outperform the state-of-the-art performance on the traffic flow prediction dataset. All models are
trained using the ADAM optimizer\cite{kingma2014adam}. The batch size of each iteration is set to 512 and 400 epochs. All experiments are implemented in TensorFlow \cite{abadi2016tensorflow} and conducted on NVIDIA Tesla K80 GPU. We used a grid search for finding the optimum deep neural network architectures which have the best performance and most efficient computational time.

The input matrix is $\vectsf{I} \in \reals^{\mathsf{s} \times \mathsf{w} \times \mathsf{k}}$, where the number of sensors is $\mathsf{s}=597$, the time window is $\mathsf{w}=6$, and there are $\mathsf{k}=3$ features, including flow, occupancy and speed. For MLP, LSTM, CNN and the proposed multi-kernel CNN-LSTM models, the input dimension is reshaped to have a appropriate dimensions, described in model details section \ref{eq::nnmodels}. For all models the data transform into range of [0, 1]. For the models without time series decomposition component, including MLP, LSTM and CNN, we transform the data into stationary data by subtracting all input values from the value at time step $\mathsf{t}$, while detrending of the models with time series decomposition components is as follows. The residual time series is stationary. To feed trends and seasonal components to a neural network, we make them stationary by subtracting each time window from its last value $\vectsf{S}^\mathsf{t}$ and $\vectsf{T}^\mathsf{t}$. To recover the output, we add the predicted value to sum of $\vectsf{S}^\mathsf{t}$ and $\vectsf{T}^\mathsf{t}$. The output matrix is $\vectsf{O} \in \reals^{\mathsf{s} \times \mathsf{h} \times \mathsf{\bar{k}}}$, where the size of horizon $\mathsf{h} = 4$ for 15-min, 30-min, 45-min and 60-min prediction in the result tables.

\subsubsection{Baseline models}
As it is illustrated in primary traffic flow prediction studies, the traffic flow patterns are similar in same hours and weekdays. The first baseline model (Ave-weekday-hourly) is to use average of traffic flow of each station as a time-table for each time $\mathsf{t}$, given a weekday. The short-term prediction for each sensor is obtained by using average values in training data. The second baseline model (current value) is to use current value of traffic flow $\vectsf{x}^\mathsf{t}$ for short-term prediction $\vectsf{x}^{\mathsf{t}+\mathsf{h}}$.

\subsubsection{State-of-the-Art Neural Network Models}\label{eq::nnmodels}
In this section, we describe the implemented neural network models. A Multi-layer perceptron (MLP) with three fully connected layers and Xavier initialization \cite{glorot2010understanding}, RELU activation function, and (500,300, 200) hidden units is used. A deep belief network (DBN) with greedy layer wise pretraining of autoencoders finds a good initialization for a fully-connected neural network. A fine tuning step for stacked auto encoder finish training. We consider 30 epochs for pretraining each layer. Fully connected Long-Short Term Memory neural network (LSTM) is capable of capturing long-term temporal patterns. However, in most of the studies, fully connected LSTM models have a shallow structure. They also are slow in convergence, although they have strong capabilities in capturing long-term patterns. The optimum LSTM neural network structure has one hidden units of size (400,200). The input is reshaped from a vector to matrix of two dimension $(\mathsf{w}, \mathsf{s} \times \mathsf{k})$. To use a convolutional neural network (CNN) for time series forecasting, the input matrix is reshaped to three dimension $(\mathsf{w}, \mathsf{s}, \mathsf{k})$. Each channel includes traffic flow, speed and occupancy. The optimum implemented deep CNN model has four layers with max-pooling and batch normalization layers. The number of filters are (16,32,64,128), the kernel is (5,5), and max-pooling layers reduce the dimension by two in each layer.Two fully connected layer connect the convolution layers to output layer. 

The (CNN-LSTM) model captures short-term and spatial patterns in CNN layers, and temporal patterns in LSTM layers. Two convolution layers are applied on all input sensors with filters (16, 32). An LSTM layer of size (300,150) follows the output of CNN model, following by a fully connected layer. The model (C-CNN-LSTM) is a clustering based CNN-LSTM, in which a multi-kernel convolution layer extract spatial, short-term patterns from time series residuals. The clusters are obtained in section \ref{eq::fhc}.

A pretraining denoising stacked auto encoder decoder is applied on each cluster of sensors to generate a robust output. Each layer is connected to drop-out layer with rate of 0.2. As the average size of clusters is nearly 10 and standard deviation of 4, described in section \ref{eq::fhc}, we used a same size of architecture for all of them with size of (40,20,10,20,40) units with fully connected layers, and RELU activation function. The pretraining is done in 60 epochs. The weights are loaded into the proposed model in the next section.

The Cluster-based CNN-LSTM with Denoising autoencoder (C-CNN-LSTM-DA) is the proposed model in section \ref{eq::dnnf} which uses clustering of time series residuals, trends, and seasonal along with denoising autoencoders and time series decomposition components for each cluster. The proposed architecture, in section \ref{eq::dnnf}, consists of 2 convolution layers with RELU and max-pooling layers with filters (32, 64). It follows by two fully connected layers, two 2-dimension convolutional LSTM for capturing long-term patterns (16, 32).

\subsection{Performance Measure}

To evaluate the performance of the proposed model, we used performance indices, Mean Absolute Error (MAE) and Root Mean Square Error (RMSE) in equation (\ref{eq::eqerror}).

\begin{align} \label{eq::eqerror}
&\mathsf{MAE}(\vectsf{y}, \vectsf{\bar{y}}) = \frac{1}{\mathsf{n}} \sum_{\mathsf{i}=1}^\mathsf{n} |\mathsf{y}_\mathsf{i} - \bar{\mathsf{y}}_\mathsf{i}|\\
&\mathsf{RMSE}(\vectsf{y}, \vectsf{\bar{y}}) = \Bigg[\frac{1}{\mathsf{n}} \sum_{\mathsf{i}=1}^\mathsf{n} (\mathsf{y}_\mathsf{i} - \bar{\mathsf{y}}_\mathsf{i})^2\Bigg]^\frac{1}{2}\\
\end{align}

Here, $\vectsf{y}$ are the real values and $\bar{\vectsf{y}}$ are the predicted values. In this paper, the prediction is for 15-min, 30-min, 45-min and 60-min time horizons.

\subsection{Performance results on testing data}
In the first analysis, we compare the performance of models for traffic flow prediction. The results are illustrated in Table \ref{tab:la_comparison}. The comparison is between all described models for prediction horizons of 15-min, 30-min, 45-min and 60-min on testing data. The two baseline models have worst performance. The performance of neural network models are much better than the baseline models. The LSTM model has a better performance than MLP, BN and CNN models, demonstrating its improved performance in time series forecasting. CNN-LSTM models are more capable for capturing short-term and long-term patterns an are comparable with LSTM. Two models, C-CNN-LSTM and C-CNN-LSTM-DA, have better performance due to explicitly separating spatial regions. The performance of C-CNN-LSTM and C-CNN-LSTM-DA is almost quite close. In next sections, we discuss that in existence of missing data, the model with denoising autoencoders have better performance.

\begin{table}[t] 
\footnotesize
\centering
\caption{Evaluation of the models for traffic flow forecasting problem.}
\label{tab:la_comparison}
\hspace{-0.5in}
\scalebox{0.8}{
\begin{tabular}{|c||c||cc|cccc|ccc|}
\cline{1-11}
\toprule
\multicolumn{2}{|l}{}&\multicolumn{2}{|l}{Baseline models}&\multicolumn{4}{|l}{State of arts neural networks}&\multicolumn{3}{|l|}{Proposed models}\\\hline
Horizon & Metric & current-value & Ave-weekday-hourly & MLP    & DBN    & LSTM    & CNN & CNN-LSTM & C-CNN-LSTM & C-CNN-LSTM-DA  \\ \hline
\midrule
\multirow{2}{*}{15 min}
& MAE  & 24 & 27.1  & 16.3   & 15.5   &  14     & 16    &  13.6  & 12.3    & \textbf{12.1} \\
& RMSE & 36 & 43.2  & 28.1   & 27     &  25     & 27.4  &  24.8  & 23.1    & \textbf{22.7}   \\ 
\cline{1-11}
\multirow{2}{*}{30 min}
& MAE  & 31 & 27.1  & 16.9   & 15.9   &  14.4   & 16.2  &  14.3  & 12.7     & \textbf{12.4} \\ 
& RMSE & 45 & 43.2  & 29     & 28   &  26.2     & 28.4  &  26.0  & 23.4     & \textbf{22.9}     \\ 
\cline{1-11}
\multirow{2}{*}{45 min}
& MAE  & 38 & 27.1  & 17.1   & 16.2   &  14.9   & 16.8  &  15  & 
12.9     & \textbf{12.8}     \\ 
& RMSE & 54 & 43.2  & 29.8   & 29     &  28.1   & 29.3  &  28.2  & 23.4     & \textbf{23.1}      \\ 
\cline{1-11}
\multirow{2}{*}{60 min}
& MAE  & 44 & 27.1  & 17.6   & 16.5   &  15.2   & 17.2 &   15.1  & \textbf{13.3}     & \textbf{13.3}    \\ 
& RMSE & 63 & 43.2  & 30.8   & 29.3   &  28.4   & 30.1  &  28.1  & 23.8     & \textbf{23.7}      \\ 
\cline{1-11}
\cline{1-11}
 \hline
\end{tabular}
}
\end{table}

\begin{figure}[h]
  \centering  
  \subfloat[Traffic flow prediction with MLP]{
    \includegraphics[height=1.3in]{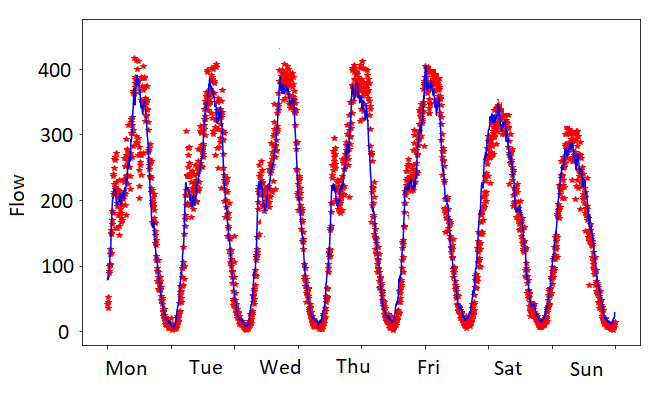}
	}
  \subfloat[Traffic flow prediction with C-CNN-LSTM-DA]
  {
    \includegraphics[height=1.3in]{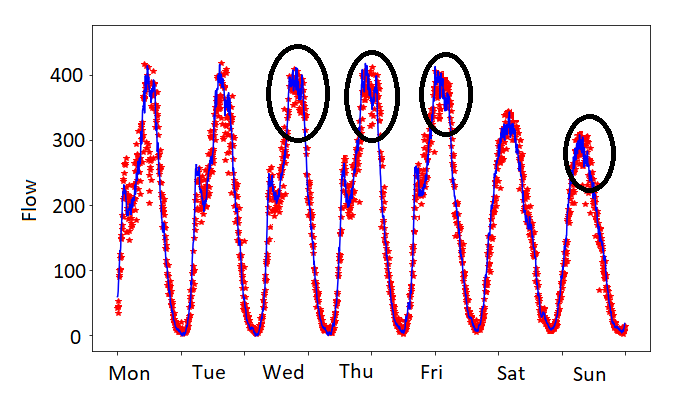}
  }
  \caption{Prediction results for traffic flow data of one sensor over one week, where the blue line is predicted, and the red is the real value. The proposed model outperforms MLP in peak hours, while they have comparable performance in off-peak hours.}\label{fig::mlp-proposed}\vspace{-0.1in}
\end{figure}

\begin{figure}[h]
\hspace*{-0.5cm}\centering
\includegraphics[width=12cm]{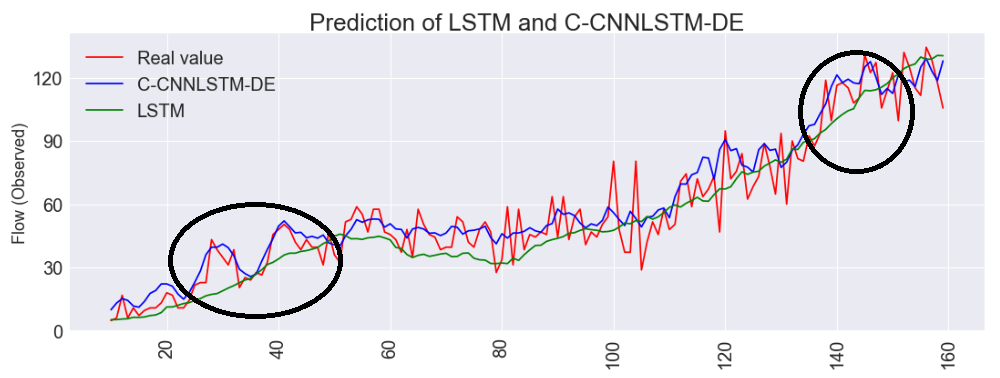}\caption{Comparison of the predictions; to illustrate the capability of the proposed model in capturing residual patterns. Some of the big fluctuations are meaningful residual patterns, and can be predicted.}\label{fig::comp-lstm-M}\vspace{-0.1in}
\end{figure}

\subsection{Performance results of peak and off-peak traffic}
Next experiment is to compare peak and off-peak hours. In peak hours, physical properties and evolution of traffic flow possibly affect congestion propagation in network. Therefor, the residuals come from the evolution of traffic and are meaningful spatial patterns. On the other hand, in off-peak hours, traffic flow is based on free speed and without congestion. Hence, flow is obtained from long-term patterns in the network. In Fig. \ref{fig::mlp-proposed}, the output of the C-CNN-LSTM-DA and MLP models are illustrated. Among neural network models, the MLP model has the worst traffic flow prediction performance, while C-CNN-LSTM-DA is the best in Table \ref{tab:la_comparison}. The C-CNN-LSTM-DA model has better performance in peak hours of the days, when there are high residuals. It shows the weakness of fully-connected neural networks for capturing residual patterns. In off-peak hours, all neural network models have comparable and good performance.

In Table \ref{tab:resid_comparison}, we compare the performance of the models for peak and off-peak traffic flow data. We select the MLP and LSTM models which only capture temporal patterns, along with the proposed model which carefully captures spatial patterns. This table compares residual-MAE and MAE on prediction values. For residual MAE, we detrend traffic flow prediction and find MAE error. For off-peak hours, LSTM has a comparable performance to the proposed model, as it simply capture long-term patterns. However, the performance of C-CNN-LSTM-DA in peak hours is highly better than LSTM model. In Fig. \ref{fig::comp-lstm-M}, we plot the comparison of LSTM and C-CNN-LSTM-DA. It is shown that C-CNN-LSTM-DA captures big residuals compared to LSTM model. While a more smooth prediction which ignores noise shows the model is not overfitted, meaningful residual patterns in spatial data need to be carefully considered in the prediction of the model.

\begin{table}[h]
\small
\centering
\caption{Performance evaluation of three models for traffic flow forecasting in peak and off-peak hours}
\label{tab:resid_comparison}
\begin{tabular}{|c||c||ccc|}
\cline{1-5}
\hline
Flow State & Metric & MLP & LSTM & C-CNN-LSTM-DA  \\ \hline
\multirow{2}{*}{Off-peak time}
& Residual MAE  & 6.1 & 4.3  & 4.4\\
& MAE & 12.3 & 11.9  & 11.8   \\ 
\cline{1-5}
\hline
\multirow{2}{*}{Peak time}
& Residual MAE  & 12.2 & 11.1  & 8.2 \\ 
& MAE & 18.3 & 16.8  & 13.8  \\ 
\cline{1-5}
 \hline
\cline{1-5}
\end{tabular} 
\end{table}

\subsection{Performance results with missing data}
We evaluate the proposed model relative to other neural network models with missing data. We randomly generated blocks of missing values in the test data. Each block $\vectsf{b}_\mathsf{j}$ is related to one randomly selected sensor $\mathsf{s}_\mathsf{j}$ at a random starting time of $\mathsf{t}_{\mathsf{s}}$, which is generated with a normal distribution with mean 2 hours and standard deviation 0.5. For each sensor one block of missing values are generated per week. We used missing data which increases error for model prediction and the real values are used for evaluation of the model. Since the missing values is only applied to one random sensor, we expect neighboring sensors to help the model to continue to predict the traffic flows well.

\begin{figure}[h]
  \centering  
  \subfloat[The increase in prediction error reduces using clustering of sensors and denoising autoencoder decoder.]{
    \includegraphics[height=1.3in]{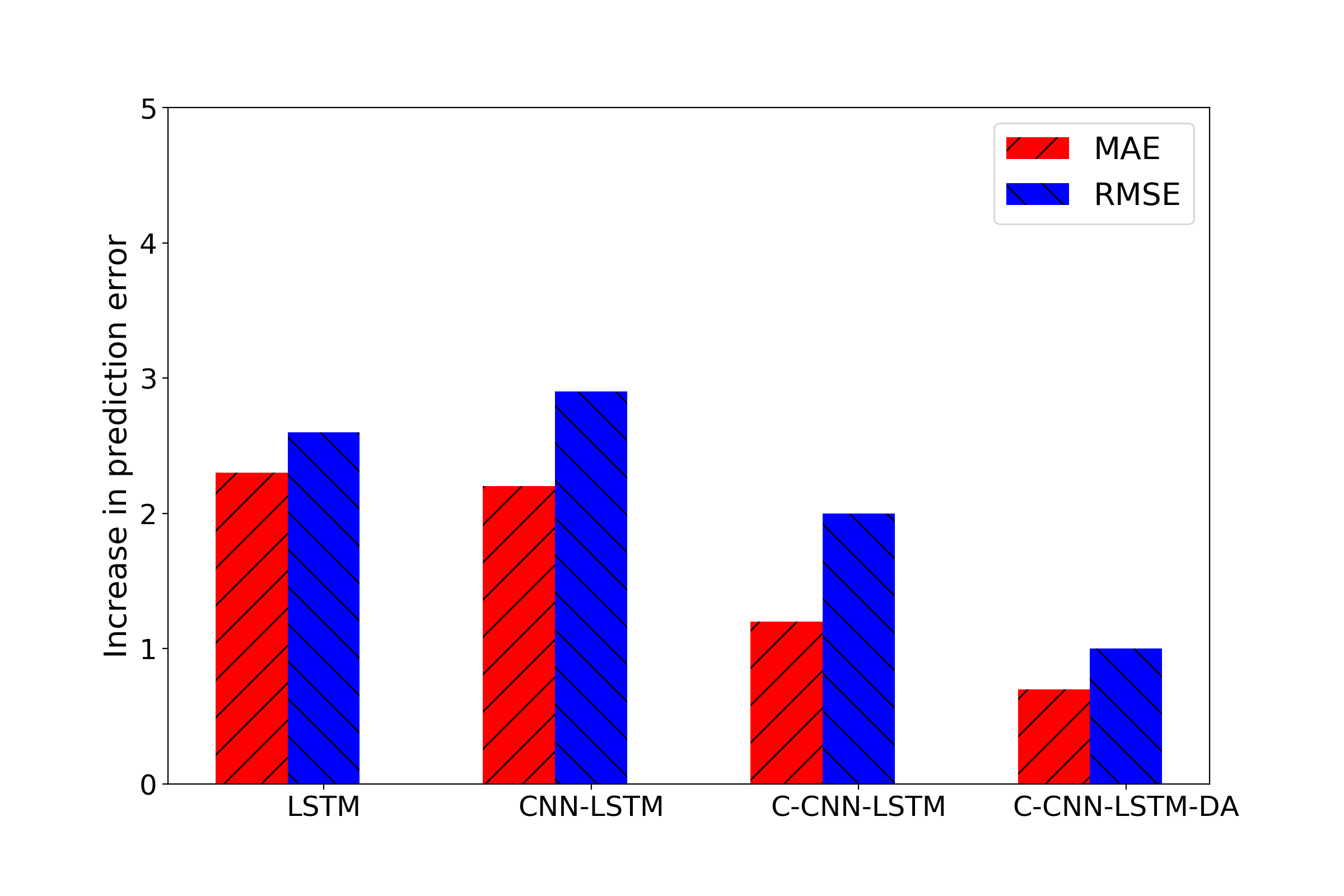}}
  \quad
  \subfloat[Comparison of prediction with random missing data]
  {
    \includegraphics[height=1.3in]{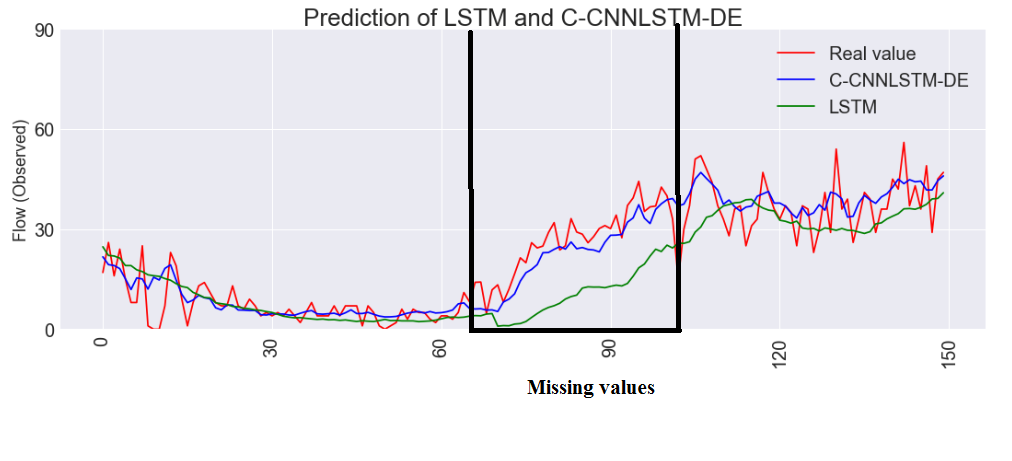}
	}
  \caption{Prediction results with random missing data}\label{fig::pred-result}\vspace{-0.1in}
\end{figure}

The results is illustrated in Table \ref{tab:miss_comparison}. To briefly describe the results, we only show the 30-min prediction in Table \ref{tab:miss_comparison}. The performance of C-CNN-LSTM-DA is better than other models in existence of missing values. In Fig. \ref{fig::pred-result}.a, we illustrate the increase in error in different models with random missing values. The figure shows a reduced increase of error in C-CNN-LSTM-DA, as the time series decomposition and denoising autoencoder components generates a more robust prediction to missing values. In the existence of missing data, the value of forecasting can be distracted far from real values in LSTM neural network, Fig. \ref{fig::pred-result}b.

\begin{table}[h]
\small
\centering
\caption{The average MAE and RMSE for the best four multi-variate time series forecasting models with randomly generated missing data.}
\label{tab:miss_comparison}
\begin{tabular}{|c||c|c|c|c||}
\cline{1-5}
\hline
Metric & LSTM    & CNN-LSTM & C-CNN-LSTM & C-CNN-LSTM-DA  \\ \hline
\hline
MAE  & 16.7      & 16.5   & 14.1   & 13.1  \\
\cline{1-5}
\hline
RMSE & 28.8      & 28.9  & 25.2   & 23.9  \\ 
\cline{1-5}
\end{tabular}
\end{table}

\section{Conclusion and future work}
This paper illustrates a new framework for spatial time series forecasting problem and its application on traffic flow data. The proposed method consists of several components. Firstly, for a time series data gathered on a network, a convolution layer does not capture network structure, because a kernel slides on spatial locations. Hence, we obtain fuzzy clusters of time series and apply a multi-kernel convolution component, in which each kernel only slides on time steps and keep the network structure of time series. Secondly, time series residuals are not a noise. They are result of interaction among neighboring time series. As an example, the similarity of time series residuals is presented in Fig. \ref{fig::ts_osf} and Fig. \ref{fig::clustering_ts}. Thus, convolution component is applied on time series residuals to extract short-term interaction among neighboring time series. In Table \ref{tab:resid_comparison}, we evaluate the prediction of time series residuals. In off-peak, the performance of the proposed model is the same as LSTM. However, the proposed model has better performance in peak hours. Table \ref{tab:la_comparison} shows the comparison of the baseline, state-of-arts neural network, and CNN-LSTM models for traffic flow prediction. The C-CNN-LSTM model uses spatial and time series decomposition, which have better performance than baseline and state-of-arts models, in Table \ref{tab:la_comparison}. One of the challenges in spatial-temporal data is to work with missing data. We illustrate the performance of using pre-trained denoising autoencoder decoder as the last component of C-CNN-LSTM-DA in Fig. \ref{fig::pred-result}. It shows the increase in error for the model with denoising autoencoder is less than other models.

This study demonstrates the effectiveness of designing new neural network architectures considering specific properties of spatial-temporal problems. Each component of the neural network is designed based on the characteristics of the extracted patterns. In the experimental results, we analyzed the spatial-temporal patterns in traffic flow data and we also illustrate the effect of each neural network components on the improvement of results. Related analyses can be constructed for other spatial-temporal problems, such as anomaly detection, missing data imputation, time series clustering and time series classification problems. In addition, different spatial-temporal problems have different physical properties or dynamical systems, which makes their patterns unique for the problems.

\bibliography{mybibfile}

\end{document}